\documentclass[letterpaper,10 pt,conference]{ieeeconf}
\IEEEoverridecommandlockouts
\usepackage{ieee_robotics}
\usepackage{dblfloatfix}
\usepackage{overpic}

\newcommand{\method}{\texttt{PreAfford}\xspace}

\title{\LARGE \bf \method: Universal Affordance-Based Pre-Grasping\\for Diverse Objects and Environments}

\author{
    Kairui Ding$^{1,2}$, Boyuan Chen$^{2}$, Ruihai Wu$^{3}$, Yuyang Li$^{4}$, Zongzheng Zhang$^{1}$, Huan-ang Gao$^{1}$, Siqi Li$^{5}$,\\ Guyue Zhou$^{1,6}$, Yixin Zhu$^{4}$, Hao Dong$^{3}$, and Hao Zhao$^{\dagger1}$\vspace{3pt}\\
    Project website: \href{https://air-discover.github.io/PreAfford}{https://air-discover.github.io/PreAfford}%
    \thanks{$^{\dagger}$ Indicates corresponding author. K. Ding, B. Chen, Z. Zhang, H. Gao, G. Zhou, and H. Zhao thank DISCOVER Robotics for providing hardware used in this research. Y. Li and Y. Zhu thank NVIDIA for providing GPUs and hardware support and are supported in part by the Beijing Nova Program.}%
    \thanks{$^{1}$ Institute for AI Industry Research (AIR), Tsinghua University. $^{2}$ Xingjian College, Tsinghua University. $^{3}$ CFCS, School of Computer Science, Peking University. $^{4}$ Institute for Artificial Intelligence, Peking University. $^{5}$ College of Control Science and Engineering, Zhejiang University. $^{6}$ School of Vehicle and Mobility, Tsinghua University.}%
    \vspace{-6pt}
}

\let\oldtwocolumn\twocolumn
\renewcommand\twocolumn[1][]{%
    \oldtwocolumn[{#1}{
    \begin{center}
        \vspace{-12pt}
        \includegraphics[width=\linewidth]{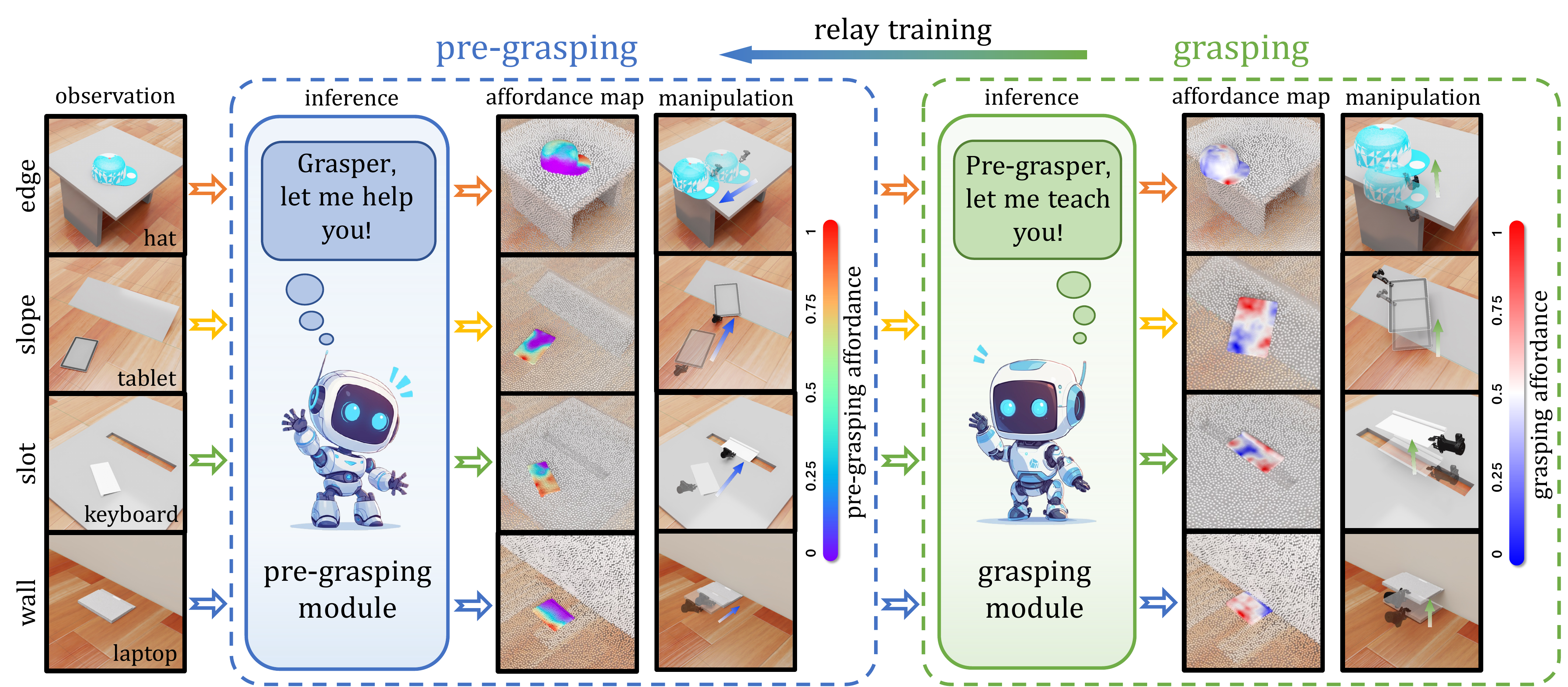}
        \captionof{figure}{\textbf{Illustration of \method, demonstrating the application of a \textit{relay training} paradigm where two synergistic modules cooperate to facilitate the manipulation of objects typically considered ungraspable.} The \textit{pre-grasping} module assesses environmental features such as edges, slopes, slots, and walls to propose strategic pre-grasping actions that enhance the likelihood of a successful grasp. Simultaneously, the \textit{grasping} module evaluates these actions and provides feedback in the form of rewards, which are used to refine and optimize the pre-grasping strategies. Two color bars represent the pre-grasping and grasping phases, respectively, with the color intensity reflecting the calculated affordance values; higher values denote more optimal interaction conditions.}
        \label{fig:teaser}
    \end{center}
    }]
}

\begin{document}
\maketitle

\begin{abstract}
Robotic manipulation with two-finger grippers is challenged by objects lacking distinct graspable features. Traditional pre-grasping methods, which typically involve repositioning objects or utilizing external aids like table edges, are limited in their adaptability across different object categories and environments. To overcome these limitations, we introduce \method, a novel pre-grasping planning framework incorporating a point-level affordance representation and a relay training approach. Our method significantly improves adaptability, allowing effective manipulation across a wide range of environments and object types. When evaluated on the ShapeNet-v2 dataset, \method not only enhances grasping success rates by 69\% but also demonstrates its practicality through successful real-world experiments. These improvements highlight \method's potential to redefine standards for robotic handling of complex manipulation tasks in diverse settings.
\end{abstract}

\section{Introduction}

\begin{table*}[t!]
    \small
    \centering
    \setlength{\tabcolsep}{3pt}
    \rowcolors{2}{LightCyan1}{}
    \caption{\textbf{Comparisons with prior pre-grasping methods.} This table offers a concise comparison of various pre-grasping studies, highlighting their key features and limitations. The last three columns evaluate each study's \textit{adaptability} across different object categories and environments, \textit{compatibility} with existing grasping pipelines via a defined procedure to bypass the pre-grasping step, and \textit{deployability} in real-world experiments without the need for custom shape representations \cite{hang2019pre} or identical environments to those used in simulators \cite{chang2010planning}. Columns A, C, and D represent adaptability, compatibility, and deployability, respectively.}
    \label{table:pre-grasping-comparison}
        \begin{tabular}{@{}cccccccc@{}}
            \toprule
            \multirow{2}{*}{\textbf{method}} & \multirow{2}{*}{\textbf{end effector}} & \multirowcell{2}{\textbf{pre-grasping}\\\textbf{manipulation}} & \multirow{2}{*}{\textbf{method}} & \multirow{2}{*}{\textbf{scenario}} & \multirow{2}{*}{\textbf{A}} & \multirow{2}{*}{\textbf{C}} & \multirow{2}{*}{\textbf{D}} \\
            & & & & & & & \\
            \midrule
            Ren \etal \cite{ren2021fast} & two-finger gripper & pushing & DRI & clustered objects & \texttimes & \checkmark & \checkmark \\
            Sun \etal \cite{sun2020learning} & spherical & rotation & DRI & cuboid in corner & \texttimes & \texttimes & \checkmark \\
            Kappler \etal \cite{kappler2010representation} & dexterous hand & pushing & data-driven approach & cuboid on table & \texttimes & \checkmark & \texttimes \\
            Chen \etal \cite{chen2023synthesizing} & dexterous hand & finger contact & learning-based framework & ungraspable cases & \checkmark & \checkmark & \texttimes \\
            Hang \etal \cite{hang2019pre} & two-finger gripper & sliding & integrated planning & thin objects on table & \texttimes & \texttimes & \texttimes \\
            Chang \etal \cite{chang2010planning} & two-finger gripper & rotation & opt. for payload & transport tasks & \texttimes & \texttimes & \texttimes \\
            Wang \etal \cite{mokhtar2022self} & two-finger gripper & pushing & DRI & clustered objects & \texttimes & \checkmark & \checkmark \\
            \textbf{Ours} & \textbf{two-finger gripper} & \textbf{pushing} & \textbf{dual-module affordance map} & \textbf{ungraspable cases} & \checkmark & \checkmark & \checkmark \\
            \bottomrule
        \end{tabular}%
\end{table*}

\begin{figure}[t!]
    \centering
    \begin{overpic}[width=\linewidth]{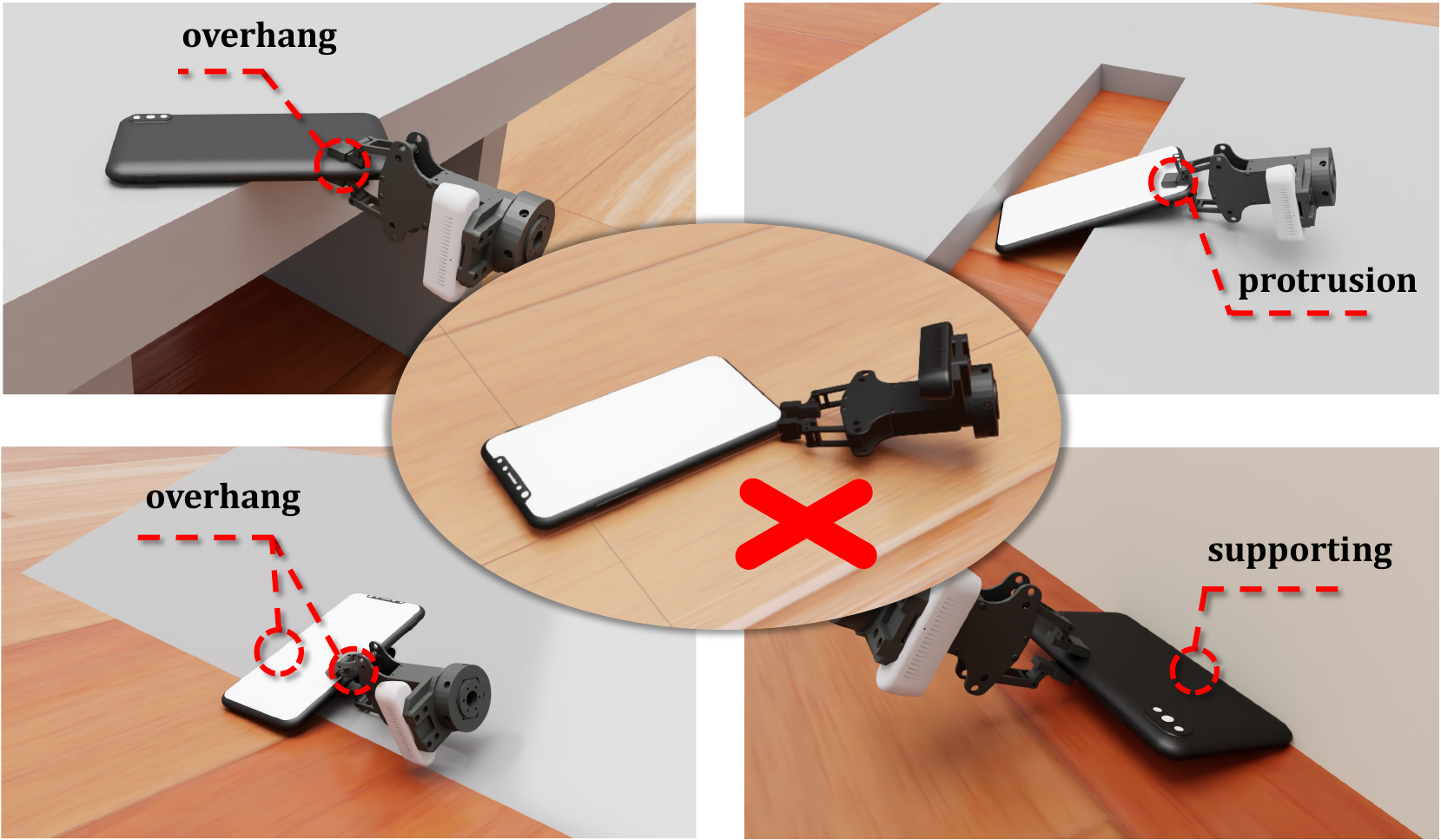}
        \put(48,38){(a)}
        \put(2,54){(b)}
        \put(94,54){(c)}
        \put(2,2){(d)}
        \put(94,2){(e)}
    \end{overpic}
    \caption{\textbf{Pre-grasping leverages environmental features to enhance graspability.} (a) An object lying flat on the floor, ungraspable in its current position. (b) Side-grasping an object that overhangs a surface. (c) Grasping an angled part protruding from a slot. (d) Grasping the middle of a phone suspended at the foot of a slope. (e) Pinning a phone against a wall and grasping it from the opposite side.}
    \label{fig:extrinsic}
\end{figure}

Consider a robotic arm equipped with a two-finger gripper attempting to grasp a mobile phone lying flat on a surface, as illustrated in \cref{fig:extrinsic}(a). The close contact between the phone and the table results in a lack of graspable features, creating an \textit{ungraspable} scenario \cite{sun2020learning}. Inspired by human object manipulation strategies, an effective approach involves repositioning the object towards a table edge, allowing the phone to hang off slightly and thus become graspable. This technique, termed \textit{pre-grasping} manipulation, is a critical preliminary step in enabling successful grasping in challenging scenarios \cite{kappler2010representation,zhou2023learning,king2013pregrasp, chang2010planning}. As depicted in \cref{fig:extrinsic}(b-e), various environmental features, also known as extrinsic dexterity \cite{lynch1999dynamic,dafle2014extrinsic}, can be leveraged to reconfigure the object into a more favorable position for grasping.

Research has shown that pre-grasping manipulations, including pushing \cite{ren2021fast}, rotating \cite{sun2020learning,chang2010planning}, and sliding \cite{hang2019pre,wujiaxi2023learning}, can significantly enhance the success rate of grasping \cite{chen2023synthesizing}. However, three key limitations remain to be addressed:
\begin{enumerate}[leftmargin=*]
    \item \textbf{Adaptability}. Existing studies primarily focus on task-specific settings, relying heavily on manually programmed criteria for pre-grasping success, limiting their ability to adapt across different object categories and environments \cite{ren2021fast,hang2019pre,sun2020learning,nguyen2016preparatory,kappler2010representation}.
    \item \textbf{Deployability}. Pre-grasping planning methods often struggle to transition to real-world applications due to their dependence on privileged information from simulators or the need for customized representations for specific experimental setups \cite{chen2023synthesizing,chang2010planning,hang2019pre}.
    \item \textbf{Compatibility}. To minimize the complexity and cost of robotic control, pre-grasping manipulations should be omitted for objects that are easy to grasp. However, mechanisms to assess the necessity of pre-grasping are often lacking in prior work \cite{ren2021fast,sun2020learning,kappler2010representation,nguyen2016preparatory,hang2019pre,wang2022adaafford,chang2010planning}.
\end{enumerate}
To date, no approach has successfully addressed adaptability, deployability, and compatibility in a unified manner.

We introduce the \method, which effectively addresses the above limitations. The adaptability of \method arises from a novel \textit{relay training} paradigm within a dual-module framework. This setup generates a robust reward function to improve the likelihood of grasping success, which is consistently evaluated by the grasping module. For deployability and real-world adaptability, \method employs a point-level affordance visual representation \cite{wu2021vat,wang2022adaafford,ning2023where2explore} that relies solely on RGB-D data. Additionally, the \method integrates a pre-grasping necessity check at the start of the inference process, enabling direct grasping of objects that are straightforward to handle, thereby enhancing compatibility and operational efficiency.

By training and testing on a large-scale offline dataset from ShapeNet-v2 \cite{chang2015shapenet} across five scenes, simulations demonstrated that \method increases the grasping success rate by 69\% for test object categories. The pre-grasping and grasping affordance maps, as shown in \cref{fig:main}, indicate that our models possess a profound understanding of object geometries and the environmental features of the system. Additionally, \method has demonstrated the ability to select appropriate pre-grasping policies in unseen, complex environments. The deployability of our framework is further confirmed through real-world experiments conducted across five different setups.

To sum up, our key contributions are:
\begin{itemize}[leftmargin=*]
    \item A novel, adaptive, and deployable pre-grasping framework, compatible with easy-to-grasp objects.
    \item A robust relay training paradigm that enhances pre-grasping manipulation strategies.
    \item Point-level affordance representation that enables detailed geometry awareness and seamless deployment.
    \item Extensive validation of \method in both simulated and real-world settings, demonstrating its efficacy.
\end{itemize}

\section{Related Work}

\subsection{Pre-Grasping Tasks}

Pre-grasping, a concept inspired by human behavior where objects are often pre-manipulated for easier handling \cite{chang2008preparatory,chang2009selection}, involves robot manipulators adjusting the pose of an object \textit{before} executing the final grasp \cite{kappler2010representation,nguyen2016preparatory,mokhtar2022self,chang2009selection,ren2021fast,li2024grasp}. This adjustment is crucial for successful interaction with objects, especially when they are in challenging positions or orientations.

We compare previous pre-grasping strategies in \cref{table:pre-grasping-comparison}. One research direction involves altering the pose of an object to a more graspable configuration, often through rotation \cite{lee2015hierarchical,chang2010planning,hang2019pre}. Another line of work focuses on pre-grasping with extrinsic dexterity \cite{lynch1999dynamic,dafle2014extrinsic,chen2023synthesizing}, which leverages environmental features to facilitate an easier grasp \cite{chen2023synthesizing,kappler2010representation,sun2020learning}. Additionally, extensive studies have addressed the issue of cluttered objects on tables, where objects overlapped in such a way that they cannot be grasped without rearrangement \cite{mokhtar2022self,ren2021fast,moll2017randomized}.

The critical element is the formulation of the reward function for pre-grasping manipulation. Defining the \textit{graspability} of an object within a specific environmental setup is challenging, leading to two main approaches for constructing the reward function. The first method involves manually programming the reward, using strategies like pre-defined goal regions \cite{hang2019pre} or specific transitions in the pose of the object and gripper \cite{sun2020learning,nguyen2016preparatory}. Although effective within their specific experimental setups, these methods often lack the flexibility to adapt to new object categories and environments. The second approach uses the output from pre-trained neural networks to estimate the \textit{graspability} at a given pose \cite{chen2023synthesizing,chang2010planning,ren2021fast}. This method typically relies on simulator-derived prior knowledge, such as signed distance functions \cite{chen2023synthesizing} or precise object and environment geometries \cite{chang2010planning}, which poses challenges for real-world application due to the necessity for a deep understanding of the scene \cite{gao2023dqs3d,gao2023semi,lee2019efficient,zhang2021invigorate}.

Furthermore, our findings indicate that many systems neglect the option to bypass the pre-grasping procedure for objects that can be directly grasped with a high confidence of success \cite{chen2023synthesizing,chang2010planning,ren2021fast}. This oversight renders the pre-grasping framework redundant for most daily graspable objects and unnecessarily increases the complexity and cost of robotic control. To date, no previous research has successfully integrated adaptability across various object-environment configurations, compatibility with directly graspable situations, and deployability in real-world scenarios in a single framework.

\subsection{Point-Level Affordance for Robotic Manipulation}

Affordance, originally defined as the action possibilities linked to an object or environment for an agent \cite{gibson1978ecological,nguyen2023open,chen2022cerberus,cui2023strap,li2022toist}, is vital in robotic manipulation. Point-level affordance learning involves creating dense affordance maps as actionable visual representations that suggest possible actions at every point on point clouds of 3D objects. Recent studies have applied point-level affordance learning to various scenarios \cite{wu2021vat,mo2021where2act,nguyen2023open,li2023gendexgrasp,ning2023where2explore,ling2024articulated,li2024unidoormanip,wu2023learningenv,zhao2022dualafford,mo2022o2o}, providing detailed and actionable information for subsequent robotic actions.

Empirical research has shown that point-level affordance has a robust geometry-aware capability, effectively generalizing across both within-category \cite{wu2023learning,wang2022adaafford} and inter-category \cite{ju2024robo,ning2023where2explore,mo2021where2act,zhao2022dualafford} scenarios. This demonstrates its adaptability to new objects and situations.

In our research, we expand the application of point-level affordance to guide pre-grasping manipulations, considering various geometries and environmental contexts. This advancement shows that point-level affordance can manage more complex scenarios with high accuracy and strong generalization potential.

\section{Methodology}

We define core concepts of our method in \cref{sec:preliminaries}, describe the overall framework in \cref{sec:framework}, elaborate on the network architectures in \cref{sec:structure}, introduce the inference procedure in \cref{sec:inference}, discuss training losses in \cref{sec:training}, and show our approach to collect data in \cref{sec:data}.

\subsection{Preliminaries}\label{sec:preliminaries}

\paragraph*{Pre-Grasping Tasks}

The primary goal of pre-grasping manipulation is to modify object poses to increase the likelihood of a successful grasp. Our approach particularly focuses on the exploitation of environmental features.

Recognizing that a single push action can effectively reposition objects, we define a pre-grasping operation \(\mathcal{P}\) as a push with an offset \(\Delta \vec{x}_1\) at the contact point \(\vec{p}_1\):
\begin{equation}
    \small
    \mathcal{P} = (\vec{p}_1, \Delta \vec{x}_1) = (x_1, y_1, z_1, \Delta x_1, \Delta y_1),
\end{equation}
where \(\vec{p}_1 = (x_1, y_1, z_1)\) and \(\Delta \vec{x}_1 = (\Delta x_1, \Delta y_1)\) denote the contact point location and horizontal displacement, respectively. By default, the direction of pushing is set to horizontal, and the gripper remains closed during this operation.

In our research, the reward for a pre-grasping action is evaluated by the increase in the graspability score yielded by the grasping module (\cref{eqn:pre-reward}). We define a pre-grasping manipulation as \textit{successful} if the score increases by more than 40\%. Two safety-critical situations are considered failure cases: (i) the object falling off the table and (ii) collision between the gripper and any obstacles such as walls.

\paragraph*{Scene-aware Grasping}

Grasping tasks involve identifying an optimal manipulation, \(\mathcal{G}\), based on object and environment information \(\mathcal{O}\). In contrast to object-centric grasping tasks, the environment plays a crucial role in scene-aware grasping, imposing constraints such as external visibility and kinematic feasibility \cite{wu2023learningenv,chen2023synthesizing}.

In our method, \(\mathcal{O}\) is represented as point clouds and a grasp \(\mathcal{G}\) is defined by six parameters:
\begin{equation}
    \small
    \mathcal{G} = (p_2, \Vec{\theta}_2) = (x_2, y_2, z_2, \alpha, \beta, \gamma),
\end{equation}
where \(p_2 = (x_2, y_2, z_2)\) denotes the contact point and \(\Vec{\theta}_2 = (\alpha, \beta, \gamma)\) the Euler angles of the grasp orientation. When the gripper closes, it moves vertically by \(\Delta z_2\) and maintains its grip for a duration \(\Delta_t\). A grasp is labeled as \textit{successful} if the object is lifted with a vertical displacement exceeding \(\theta_v\) and the rotation \(\sqrt{\alpha^2 + \beta^2 + \gamma^2}\) remains within \(\theta_{r2}\).

\subsection{Overview of the \method}\label{sec:framework}

\begin{figure*}[t!]
    \centering
    \includegraphics[width=\linewidth]{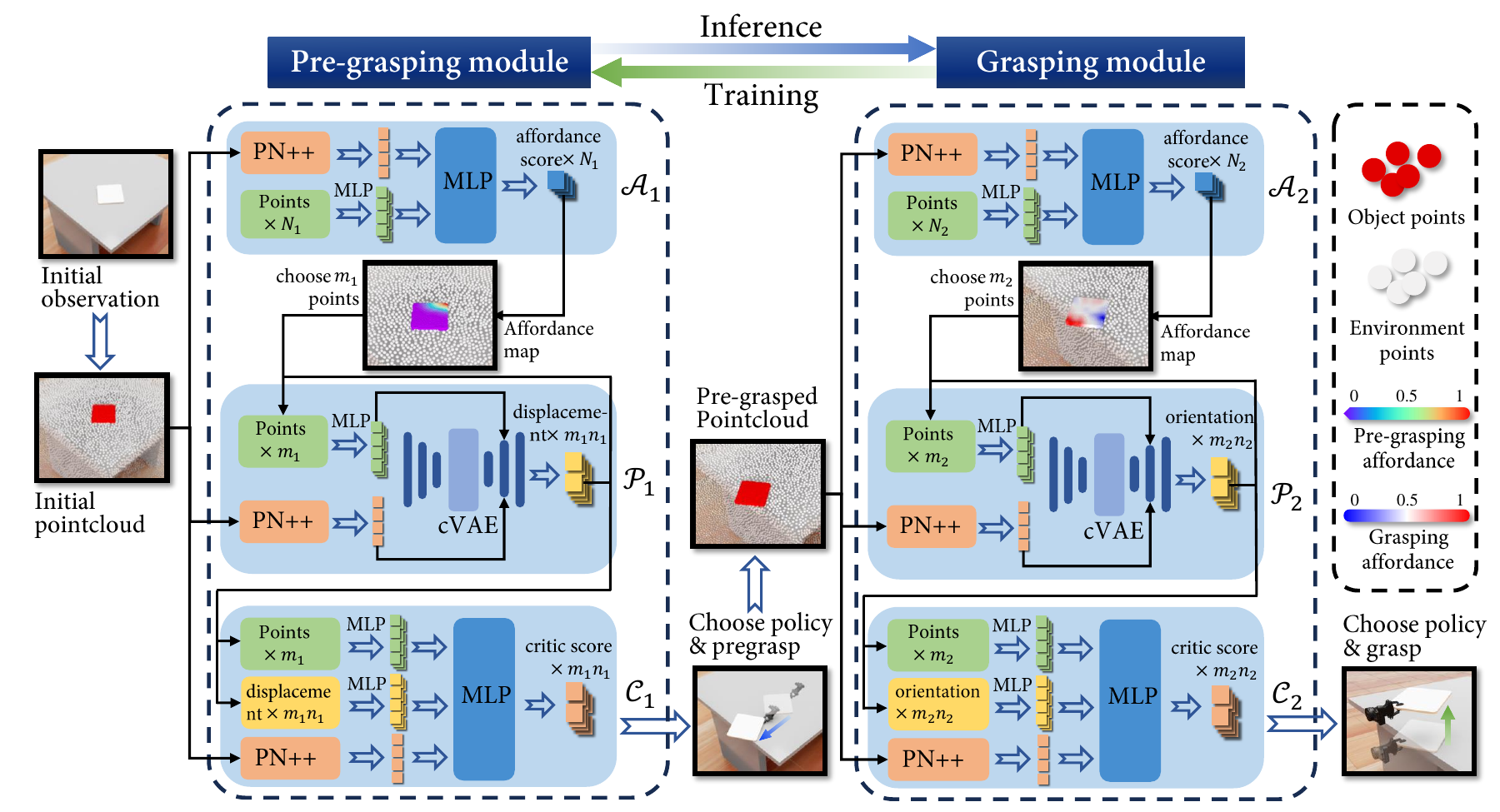}
    \caption{\textbf{The framework of \method.} The framework consists of two main modules, each incorporating three networks: an affordance network, a proposal network, and a critic network. These networks respectively handle tasks of choosing the contact point, generating a proposal, and evaluating the proposal. PointNet++ (PN++) and MLP are employed to process point clouds and facilitate decision-making. During the inference phase, both modules collaborate to develop strategies for pre-grasping and grasping. In contrast, during the training phase, the grasping module generates rewards for training the pre-grasping module, a process we refer to as \textit{relay}.}
    \label{fig:framework}
\end{figure*}

\Cref{fig:framework} illustrates the \method framework. We divide the task into two main phases: pre-grasping and grasping, which are managed by the pre-grasping module and the grasping module, respectively. Within each module, three specialized neural networks, inspired by the \textit{Where2Act} structure \cite{mo2021where2act}, are employed: an affordance network \(\mathcal{A}\), a proposal network \(\mathcal{P}\), and a critic network \(\mathcal{C}\).

Training and inference processes operate in opposite directions, a concept we refer to as \textit{relay}. The offline training dataset is collected from simulations. A grasping network, which judges the likelihood of a grasp's success, is trained first to generate labels for training the pre-grasping module. Subsequently, the pre-grasping network is trained using both the simulation data and the outputs from the grasping network. During the prediction phase, the two modules form a closed-loop control system. The pre-grasping module adjusts the object's pose until it is deemed suitable for the grasping module to apply a grasp.

\subsection{Module Structure}\label{sec:structure}

\paragraph*{Feature Extractors}

Each network within the modules independently extracts features through its perception module. Object and environment point clouds are encoded into feature vectors \( f_{o} \in \mathbb{R}^{160} \) using a PointNet++ module equipped with a segmentation head \cite{qi2017pointnet++}. Furthermore, various Multilayer Perceptron (MLP) networks are employed to process additional input features. These include encoding the contact point \( p_i \) into \( f_{p_i} \in \mathbb{R}^{32} \), the gripper displacement \( \Delta \Vec{x}_1 \) into \( f_{M_1} \in \mathbb{R}^{32} \), and the gripper orientation \( \Vec{\theta}_2 \) into \( f_{M_2} \in \mathbb{R}^{32} \).

\paragraph*{Affordance Network}

The Affordance Networks, \( \mathcal{A}_1 \) and \( \mathcal{A}_2 \), compute an affordance score \( \mathcal{A}_i(p_i|\mathcal{O}_i) \) within the range [0,1] for each specific contact point \( p_i \). The score from \( \mathcal{A}_1 \) evaluates the suitability of contact points for pre-grasping, while \( \mathcal{A}_2 \) assesses the likelihood of successful grasping. These networks are implemented using Multilayer Perceptron (MLP) models that take feature vectors \( f_{o} \) and \( f_{p_i} \) as inputs. The collective affordance scores are then used to create an affordance map, as illustrated in \cref{fig:manipulation}.

\paragraph*{Proposal Network}

The Proposal Network \( \mathcal{P}_1 \) is designed to output the gripper displacement \( \Delta \vec{x}_1 \) at the specified contact point \( p_1 \). It utilizes a conditional Variational Autoencoder (cVAE) architecture \cite{cortes2015advances} with 32 hidden dimensions. Within this structure, the encoder maps the feature vectors \( f_{o} \), \( f_{p_1} \), and \( f_{M_1} \) into a Gaussian noise vector \( z \in \mathbb{R}^{32} \), which the decoder then reconstructs. Similarly, \( \mathcal{P}_2 \) operates with the same architecture but is focused on receiving and reconstructing the gripper orientation \( \vec{\theta}_2 \).

\paragraph*{Critic Network}

A network \( \mathcal{C}_1 \) evaluates the efficacy of a pre-grasping operation by assigning a score \( \mathcal{C}_1((p_1, \Delta \vec{x}_1) \mid \mathcal{O}_1) \in \mathbb{R} \). This score considers the pair \( (p_1, \Delta \vec{x}_1) \) proposed by \( \mathcal{A}_1 \) and \( \mathcal{P}_1 \). Similarly, \( \mathcal{C}_2 \) assesses the likelihood of success for a grasping operation, providing a score \( \mathcal{C}_2(p_2, \vec{\theta}_2 \mid \mathcal{O}_2) \in [0,1] \). Both critic networks utilize an MLP to process the input features \( f_{o} \), \( f_{p_i} \), and \( f_{M_i} \).

\subsection{Inference}\label{sec:inference}

The inference pipeline of the proposed method includes four stages, which are described respectively below.

\paragraph*{Pre-grasping Necessity Check}\label{sec:necessity}

Prior to the pre-grasping step, a necessity check is performed to determine whether this step can be skipped. This decision is based on an evaluation of \textit{graspability}, where the network \( \mathcal{C}_2 \) assesses manipulation proposals generated by \( \mathcal{A}_2 \) and \( \mathcal{P}_2 \). The success rate \( \hat{c}_2 \) is calculated by averaging the scores according to the following equation:
\begin{equation}
    \small
    \hat{c}_2 = \frac{1}{n_2 m_2} \sum_{j=1}^{n_2} \sum_{k=1}^{m_2} \mathcal{C}_2(p_{2}^j, \mathcal{P}_2 (p_{2}^j, z^k) \mid \mathcal{O}_2)\label{eqn:c_2}.
\end{equation}
If \( \hat{c}_2 \) exceeds the threshold \( \theta_g \), the object will be grasped directly according to the proposal with the highest score.

\paragraph*{Pre-Grasping Manipulation Inference and Implementation}
During the inference stage, the affordance network \( \mathcal{A}_1 \) assesses each potential contact point's affordance value. The top \( n_1 \) points, denoted as \( p_1^j \) (for \( j = 1, 2, \ldots, n_1 \)), are selected based on the highest scores. Subsequently, for each selected point \( p_1^j \), the proposal network \( \mathcal{P}_1 \) generates \( m_1 \) potential pre-grasping manipulations \( \Delta \vec{x}_1^{jk} \) (where \( k = 1, 2, \ldots, m_1 \)), each paired with a randomly generated Gaussian noise vector \( z_1^{jk} \).

The critic network \( \mathcal{C}_1 \) then evaluates these manipulations to select the optimal pair \( (p_1^{j^*}, \Delta \vec{x}_1^{j^*k^*}) \) for execution, based on the following criteria:
\begin{equation}
    \resizebox{.91\linewidth}{!}{$%
        p_1^{j}(\mathcal{O}_1) = \underset{p_1}{\mathrm{argmax}}^{(n_1)} \mathcal{A}_1(p_1 \mid \mathcal{O}_1),\quad{} (j^*, k^*) = \underset{j,k}{\mathrm{argmax}} \, \mathcal{C}_1(p_1^j, \mathcal{P}_1(p_1^j, z_1^{jk}) \mid \mathcal{O}_1).
    $}
\end{equation}
This procedure modifies the object's pose within the environment, resulting in a new point cloud \( \mathcal{O}_2 \), which serves as the input for the subsequent grasping module.

\paragraph*{Grasping Manipulation Inference and Implementation}

Following the pre-grasping step, as discussed in \cref{sec:necessity}, the object is ready to be grasped. The final selection of the manipulation technique is determined by the highest critic score. This score is produced by evaluating the combined output of the affordance network \( \mathcal{A}_2 \) and the proposal network \( \mathcal{P}_2 \). The manipulation with the highest score is then executed to grasp the object. This ensures that the chosen approach is optimal based on the current state of the object within the environment.

\paragraph*{Closed-loop Control}

The inference pipeline of \method forms a closed-loop control system, enabling iterative refinement of the object's position to achieve a configuration suitable for grasping. This optional feature triggers additional manipulations when an initial pre-grasping action fails to render the object graspable. The closed-loop process operates as outlined below:
\begin{enumerate}[leftmargin=*]
    \item The grasping module initially assesses the expected success rate of the grasp, denoted by \( \hat{c}_2 \).
    \item If \( \hat{c}_2 \) is below a predefined threshold \( \theta \), the pre-grasping module executes a manipulation to reposition the object.
    \item The system then returns to step 1 to reassess \( \hat{c}_2 \) under the new object configuration.
    \item This cycle (steps 2 and 3) repeats until \( \hat{c}_2 \) exceeds the threshold, indicating that the object is in a graspable position, at which point an attempt to grasp is made.
\end{enumerate}

This feedback mechanism ensures that the system dynamically adapts to environmental changes, enhancing the likelihood of successful grasping through iterative adjustments.

\subsection{Training and Losses}\label{sec:training}

\paragraph*{Critic Loss}

The loss function for the critic network $\mathcal{C}_2$ is formulated based on the binary outcome (success or failure) of the grasping manipulation, denoted by $r$. The loss is defined as follows:
\begin{equation}
    \small
    \mathcal{L}_{\mathcal{C}_2} = r\log(\mathcal{C}_2(p_2, \Vec{\theta}_2)) + (1-r)\log(1-\mathcal{C}_2(p_2, \Vec{\theta}_2)).
\end{equation}
This binary cross-entropy loss measures the performance of $\mathcal{C}_2$ in predicting the success of grasping actions.

For $\mathcal{C}_1$, the loss is computed based on how effectively the pre-grasping manipulation improves the likelihood of a successful grasp. The success rates before and after the manipulation are evaluated by the grasping module, yielding $\hat{c}_{2}^{\rm before}$ and $\hat{c}_{2}^{\rm after}$, respectively. To encourage stable and safe manipulation actions, we apply penalties to the improvement in grasping success likelihood ($\hat{c}_{2}^{\rm after} - \hat{c}_{2}^{\rm before}$) based:
\begin{itemize}[leftmargin=*]
    \item \textbf{Displacement Penalty:} Defined as $p_d = \exp(-|\Delta \mathbf{x}_{\rm go}| / a)$, where $\Delta \mathbf{x}_{\rm go}$ represents the relative displacement between the grippers and the object, and $a$ is a scaling coefficient.
    \item \textbf{Rotation Penalty:} Defined as $p_r = \exp(-\sqrt{\alpha^2 + \beta^2 + \gamma^2} / b)$, where $\alpha$, $\beta$, and $\gamma$ are the Euler angles describing the rotation, and $b$ is a scaling coefficient.
    \item \textbf{Safety Penalty:} $p_s$ is initially set to $1$, but is reduced to $0$ in scenarios such as the object falling off the table, or collisions involving the gripper with walls or slopes.
\end{itemize}

The total penalty term $p$ is computed as the product of $p_d$, $p_r$, and $p_s$. The loss for $\mathcal{C}_1$ incorporates these penalties and is defined using the $l_1$ loss metric:
\begin{equation}
    \small
    \mathcal{L}_{\mathcal{C}_1} = \left| \mathcal{C}_1(p_1, \Delta \vec{x}_1|\mathcal{O}_1) - p \cdot (\hat{c}_{2}^{\rm after} - \hat{c}_{2}^{\rm before}) \right|.
    \label{eqn:pre-reward}
\end{equation}

\paragraph*{Proposal Loss}

The networks $\mathcal{P}_1$ and $\mathcal{P}_2$ are implemented as cVAEs, designed to generate appropriate actions for manipulation tasks. The training process exclusively uses data from \textit{successful} pre-grasping and grasping manipulations to ensure the generation of effective actions. The loss function for these networks is a combination of geometric and Kullback-Leibler (KL) divergence losses. The geometric loss, $\mathcal{L}_{\rm geo}$, quantifies the discrepancy between the reconstructed manipulation, $M_i$, and the ground truth, $\hat{M}_i$. The KL divergence, $D_{\rm KL}$, assesses the deviation of the hidden layer distribution from a standard normal distribution. The total loss for each proposal network is expressed as: $\mathcal{L}_{\mathcal{P}_i} = \mathcal{L}_{\rm geo}(M_i; \hat{M}_i) + D_{\rm KL}(z(p_i, \hat{M}_i, \mathcal{O}_i) \parallel \mathcal{N}(0,1))$.

\begin{figure}[b!]
    \centering
    \begin{subfigure}[b]{0.492\linewidth}
        \includegraphics[width=\linewidth]{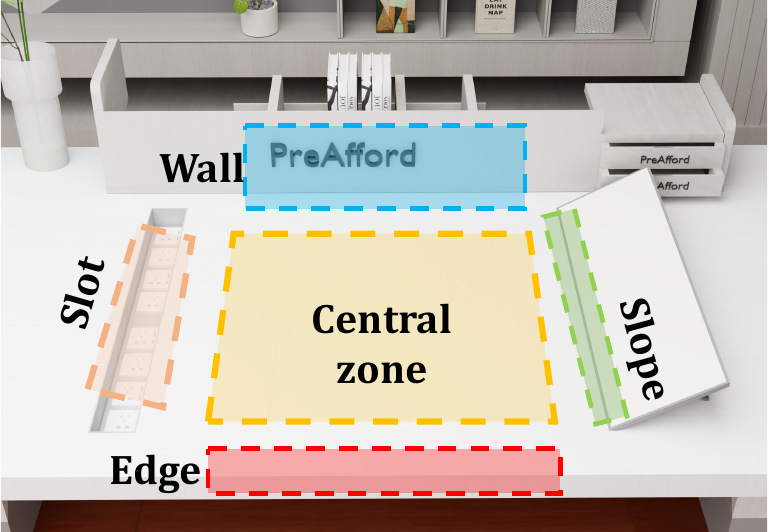}
        \caption{A complex environment}
    \end{subfigure}%
    \hfill%
    \begin{subfigure}[b]{0.5\linewidth}
        \includegraphics[width=\linewidth]{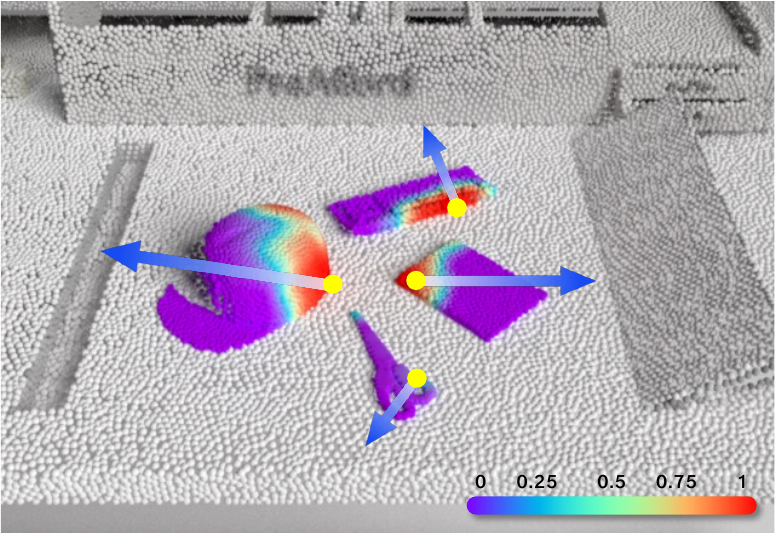}
        \caption{Affordance heatmap}
    \end{subfigure}%
    \caption{\textbf{Multi-feature scenario:} \method effectively addresses scenarios where multiple environmental features are present simultaneously.}
    \label{fig:multi}
\end{figure}

\begin{figure*}[b!]
    \centering
    \includegraphics[width=\linewidth]{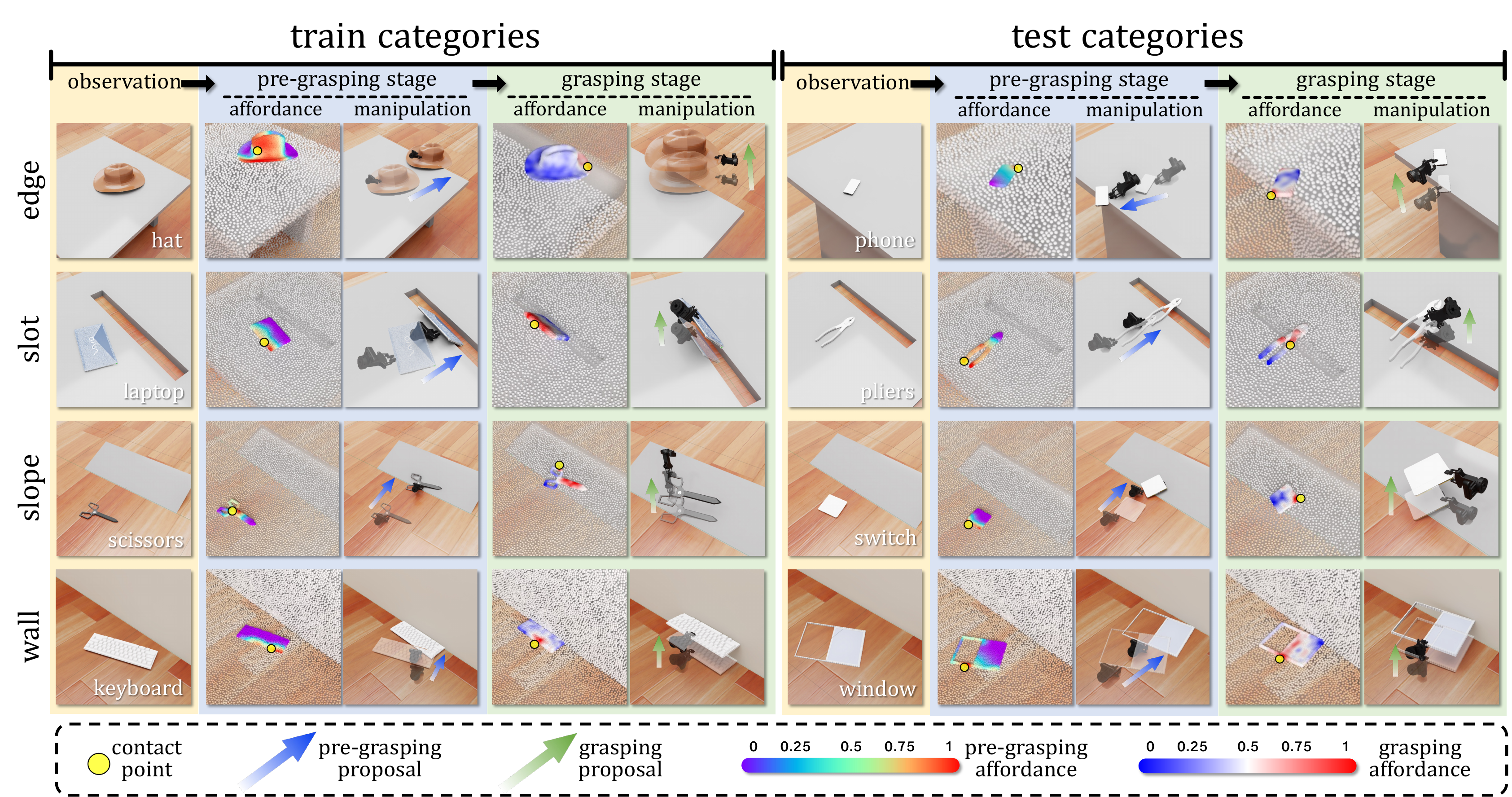}
    \caption{\textbf{Qualitative Results.} We demonstrate pre-grasping manipulation on training and testing categories in four scenarios—edge, slot, slope, and wall. Affordance maps highlight effective interaction areas, showing \method's capability to devise suitable pre-grasping and grasping strategies for various object categories and scenes, including both seen and unseen objects.}
    \label{fig:main}
\end{figure*}

\paragraph*{Affordance Loss}

The affordance score is a critical measure in robotic manipulation that assesses the suitability of contact points, guiding the selection process during action proposals. It evaluates the efficacy of actions generated by the proposal networks $\mathcal{P}_i$. These scores are calculated offline using the critic networks, which gauge the performance of multiple actions under varying conditions. The affordance network is trained subsequent to the other networks in each module. The dataset for training is labeled with the average critic scores $\hat{a}_{p_i}$, computed from $n_i$ actions generated by $\mathcal{P}_i$ using different Gaussian noise vectors $z_j$ ($j=1,2,\ldots,n_i$). The affordance loss, $\mathcal{L}_{\mathcal{A}_i}$, measures the absolute difference between the predicted and the averaged critic scores:
\begin{equation}
    \resizebox{.91\linewidth}{!}{$%
        \hat{a}_{p_i} = \frac{1}{n_i} \sum^{n_i}_{j=1} \mathcal{C}_i(p_i, \mathcal{P}_i(p_i, z_j) \mid \mathcal{O}_i),\quad{} \mathcal{L}_{\mathcal{A}_i} = \left| \mathcal{A}_i(p_i \mid \mathcal{O}_i) - \hat{a}_{p_i} \right|.
    $}
\end{equation}

\subsection{Data Collection}\label{sec:data}

We used the Sapien simulation platform \cite{xiang2020sapien} to collect data for robotic pre-grasping and grasping tasks. Our dataset, drawn from ShapeNet-v2, includes 5 hard-to-grasp and 5 easy-to-grasp object categories, each with more than 10 unique shapes.

The INSPIRE-ROBOTS EG2-4C gripper model was used in grasping trials across four simulated scenes. Objects were randomly positioned, sometimes on environmental features like tables or ledges. Grasping points were randomly chosen from the object's surface, with gripper orientations determined by a hemisphere above each point's tangent plane.

The pre-grasping dataset concentrated on hard-to-grasp categories. Objects were placed at random distances from features, and the gripper performed a pushing action with a normally distributed displacement at random surface points. To improve data collection efficiency by about 42\%, the displacement direction towards features was preset in 30\% of trials based on domain knowledge, with interaction angles modeled by a Gaussian distribution.

Both datasets positioned the camera 3–5 meters away, oriented towards the object to simulate realistic robotic vision constraints. Each dataset included 10,000 successful and 30,000 unsuccessful interactions. The data-collection process takes about 8 days on 256 AMD EPYC 7742 processors.

\section{Experiments}

\subsection{Simulation Settings and Datasets}

We tested \method on 5 seen (hat, laptop, scissors, keyboard, and tablet) and 4 unseen (phone, pliers, switch, and window) categories of hard-to-grasp objects across 5 scenes. These scenes include four with a single environmental feature (edge, wall, slope, and slot) and one novel scene (shown in \cref{fig:multi}) that contains all four features simultaneously, which is constructed to test adaptability to complex and unseen environments. We conducted 1,000 tests on each object-environment pair and calculated the mean success ratio.

\subsection{Simulation Evaluation Metrics and Baselines}

To assess pre-grasping proposals, we measure the increase in grasping success rate following a pre-grasping manipulation. After each manipulation, a grasping proposal is generated and tested 1,000 times per object-environment pair for robust success rates. The effectiveness of the pre-grasping manipulation is evaluated by comparing the enhanced success rate against scenarios without pre-grasping.

We compare against four baselines:
(i) W/o pre-grasping: direct grasping without pre-grasping actions;
(ii) Random-direction Push: the contact point is set by our pre-grasping module, but displacement is random;
(iii) Center-point Push: displacement is set by our module, but the contact point is at the object's geometric center;
(iv) Ours w/o closed-loop: an ablation study eliminating closed-loop control from our method.
Additionally, our method's compatibility is demonstrated across 5 training and 4 testing categories of easy-to-grasp objects.

\begin{figure*}[b!]
    \centering
    \begin{subfigure}[b]{0.25\linewidth}
        \includegraphics[width=\linewidth]{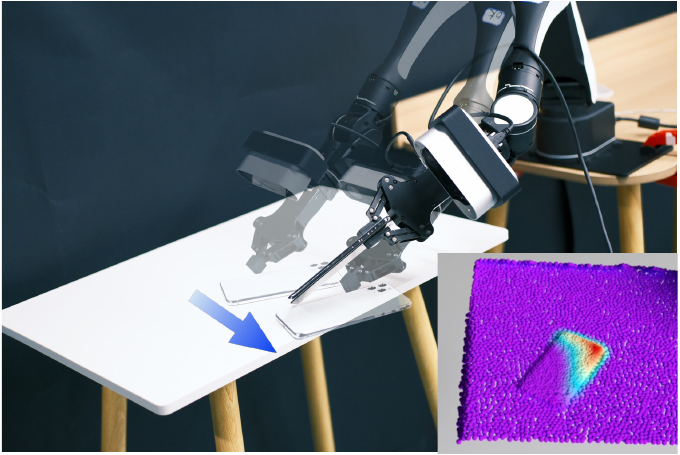}
        \caption{move a tablet to table edge}
    \end{subfigure}%
    \begin{subfigure}[b]{0.25\linewidth}
        \includegraphics[width=\linewidth]{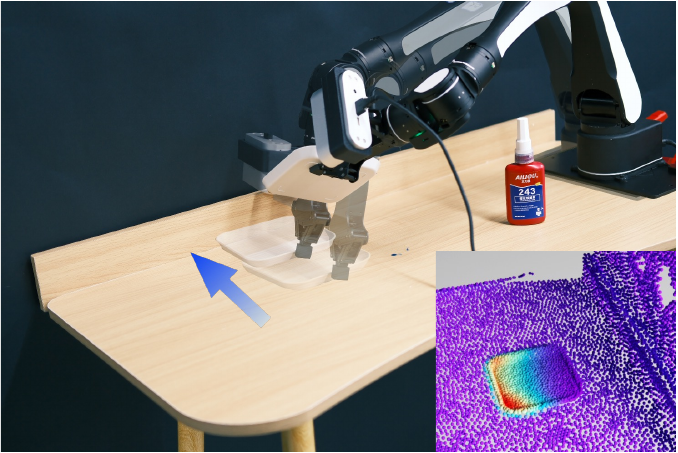}
        \caption{push a plate towards a wall}
    \end{subfigure}%
    \begin{subfigure}[b]{0.25\linewidth}
        \includegraphics[width=\linewidth]{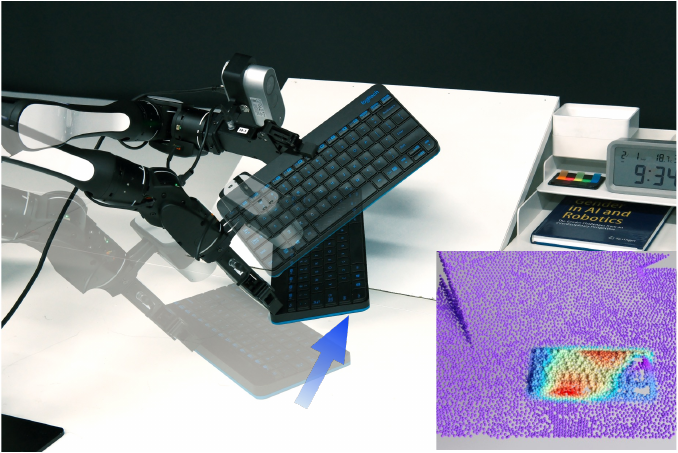}
        \caption{push a keyboard up a slope}
    \end{subfigure}%
    \begin{subfigure}[b]{0.25\linewidth}
        \includegraphics[width=\linewidth]{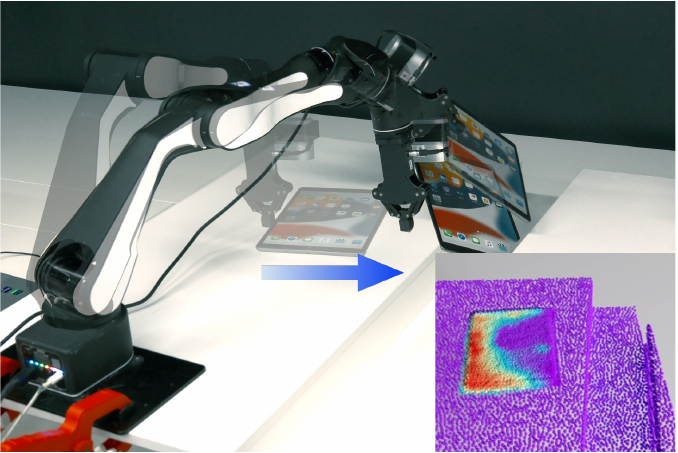}
        \caption{slide a tablet into a slot}
    \end{subfigure}%
    \caption{\textbf{Real-world pre-grasping manipulations with affordance maps.} Red areas in the maps indicate optimal pushing locations. Point clouds are captured by Femto Bolt.}
    \label{fig:manipulation}
\end{figure*}

\subsection{Simulation Results and Analysis}

\paragraph*{Efficacy and Adaptability}

\Cref{fig:main} illustrates the pre-grasping and grasping affordance maps predicted by our networks alongside the displacement suggested by $\mathcal{P}_1$. Key findings include:
(i) \textbf{Environmental awareness}: Our network effectively recognizes environmental features and proposes suitable pre-grasping actions, directing movements towards beneficial landscapes.
(ii) \textbf{Dynamics awareness}: It also displays an understanding of object dynamics, favoring pushes that align with the center of mass to reduce rotation.

The adaptability of \method is showcased by effective pre-grasping proposals across varied scenes and objects, as seen in \cref{fig:multi}, which handles complex environments with multiple features adeptly.

\Cref{table:pre-grasping-baseline} quantitatively validates \method, highlighting its superiority over direct grasping in challenging situations. Both Random-direction Push and Center-point Push strategies underline the importance of informed direction and contact point selection. Additionally, the ablation study without closed-loop control emphasizes the critical role of this feature in enhancing success rates.

\begin{table}[htb!]
    \small
    \centering
    \setlength{\tabcolsep}{3pt}
    \caption{\textbf{Success rates of grasping manipulation in percentage.} Pre-grasping increases grasping success rates by 52.9\%. A closed-loop strategy further enhances this improvement by 16.4\% across all categories.}
    \label{table:pre-grasping-baseline}
    \resizebox{\linewidth}{!}{%
        \begin{tabular}{@{}ccccccccccccc@{}}
            \toprule
            \multirow{2}{*}{\textbf{Setting}} & \multicolumn{6}{c}{\textbf{Train object categories}} & \multicolumn{6}{c}{\textbf{Test object categories}} \\ 
             \cmidrule(lr){2-7} \cmidrule(lr){8-13}
             & Edge & Wall & Slope & Slot & Multi & Avg. & Edge & Wall & Slope & Slot & Multi & Avg. \\ 
            \midrule
            W/o pre-grasping & 2.3 & 3.8 & 4.3 & 3.4 & 4.0 & 3.6 & 6.1 & 2.3 & 2.9 & 5.7 & 6.0 & 4.6\\
            Random-direction Push & 21.6 & 10.3 & 6.4 & 16.8 & 18.1 & 14.6 & 24.9 & 17.2 & 12.1 & 18.4 & 23.0 & 19.1\\
            Center-point Push & 32.5 & 23.7 & 40.5 & 39.2 & 39.0 & 35.0 & 25.1 & 17.4 & 28.0 & 30.2 & 21.5 & 24.4 \\
            Ours w/o closed-loop & 67.2 & 41.5 & 58.3 & 76.9 & 63.6 & 61.5 & 56.4 & 37.3 & 62.6 & 75.8 & 55.4 & 57.5\\
            \textbf{Ours} & \textbf{81.4} & \textbf{43.4} & \textbf{73.1} & \textbf{83.5} &  \textbf{74.1} & \textbf{71.1} & \textbf{83.7} & \textbf{47.6} & \textbf{80.5} & \textbf{83.0} & \textbf{74.6} & \textbf{73.9}\\
            \bottomrule
        \end{tabular}%
    }%
\end{table}

\paragraph*{Compatibility}

As detailed in \cref{sec:inference}, cases where the estimated grasping success likelihood $\hat{c}_2$ falls below a threshold $\theta_g$ prompt a pre-grasping manipulation. The selection of $\theta_g$ is pivotal for ensuring compatibility with graspable objects and for activating necessary pre-grasping manipulations in ungraspable scenarios. Our tests in the Multi scene demonstrate that a $\theta_g$ of 0.8 strikes an optimal balance, as evidenced in \Cref{table:compatibility}.

\begin{table}[htb!]
    \small
    \centering
    \setlength{\tabcolsep}{3pt}
    \caption{\textbf{Rate of performing a direct grasping on both graspable and ungraspable objects.} While $\theta_g=0.8$, most ungraspable objects would be pre-grasped, while graspable objects not. G: graspable; U: ungraspable.} 
    \label{table:compatibility}
        \begin{tabular}{@{}ccccc@{}}
            \toprule
            \multirow{2}{*}{\textbf{Metric}} & \multicolumn{2}{c}{\textbf{Train}} & \multicolumn{2}{c}{\textbf{Test}} \\ 
            \cmidrule(lr){2-3} \cmidrule(lr){4-5}
             & G & U & G & U \\ 
            \midrule
            pre-grasping rate & 14.7 & 84.1 & 23.4 & 77.5 \\
            success rate & 83.7 & 74.1 & 80.0 & 74.6 \\
            \bottomrule
        \end{tabular}%
\end{table}

\subsection{Real-world Experiment Setup}

We implemented our algorithm on the AIRBOT Play robotic arm, a compact six-degree-of-freedom manipulator. To capture RGB-D data, we used an ORBBEC Femto Bolt camera. The system's end effector is an INSPIRE-ROBOTS EG2-4C gripper.

\begin{figure}[t!]
    \centering
    \begin{minipage}{0.66\linewidth}
        \begin{subfigure}[b]{\linewidth}
            \includegraphics[width=\linewidth]{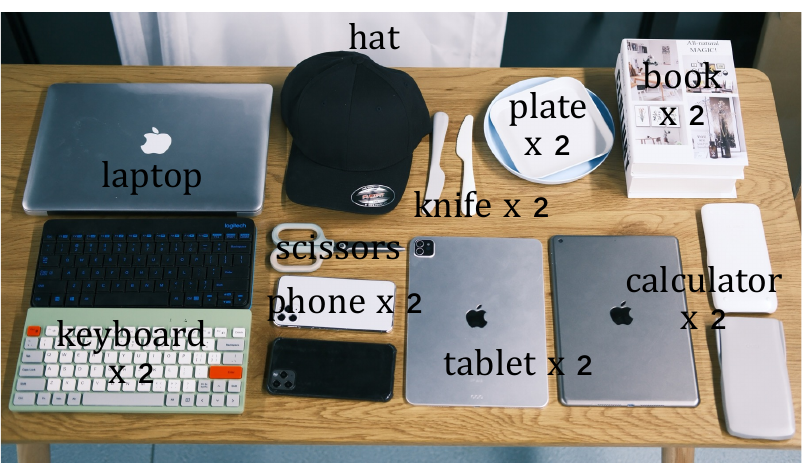}
            \caption{Object categories for testing.}
        \end{subfigure}
        \\
        \begin{subfigure}[b]{\linewidth}
            \includegraphics[width=\linewidth]{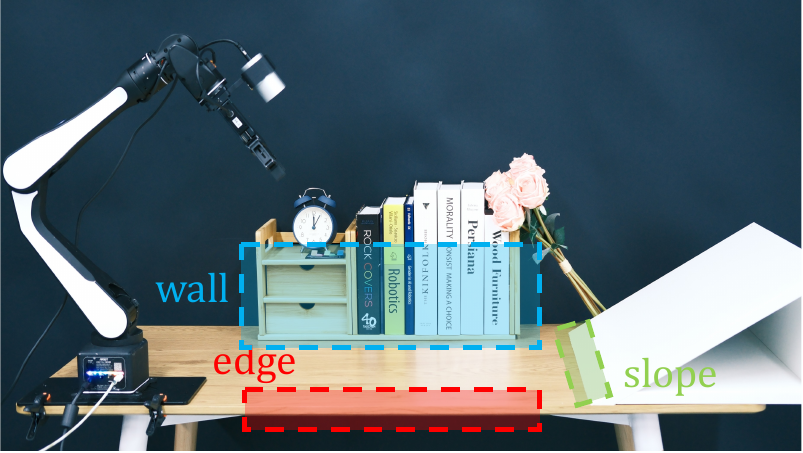}
            \caption{Multi-scene setup with three environmental features. Hardware includes an AIRBOT Play robotic arm, an INSPIRE-ROBOTS gripper, and a Femto Bolt RGB-D camera.}
        \end{subfigure}
    \end{minipage}%
    \begin{minipage}{0.34\linewidth}
        \begin{subfigure}[b]{\linewidth}
            \includegraphics[width=\linewidth]{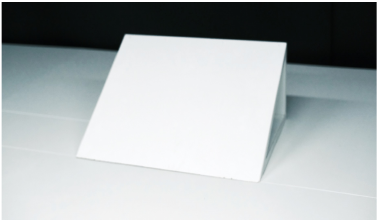}
            \caption{Slope.}
        \end{subfigure}
        \\
        \begin{subfigure}[b]{\linewidth}
            \includegraphics[width=\linewidth]{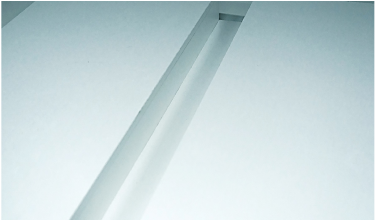}
            \caption{Slot.}
        \end{subfigure}
        \\
        \begin{subfigure}[b]{\linewidth}
            \includegraphics[width=\linewidth]{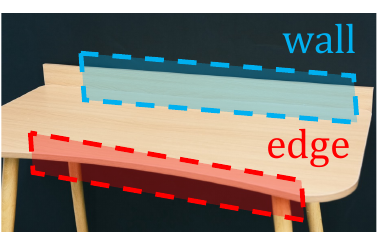}
            \caption{Wall with edge.}
        \end{subfigure}
        \\
        \begin{subfigure}[b]{\linewidth}
            \includegraphics[width=\linewidth]{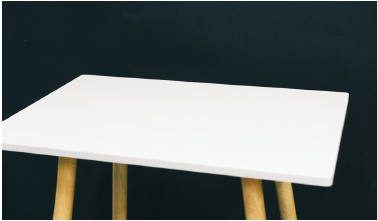}
            \caption{Edge.}
        \end{subfigure}
    \end{minipage}
    \caption{\textbf{Real-world experiment setups.}}
    \label{fig:setups}
\end{figure}

The experimental arrangement included ten object categories, split into five known and five unknown categories during training, as depicted in \cref{fig:setups}(a). To assess the adaptability of \method, we designed five distinct scenes incorporating various environmental features: edges, walls, slopes, slots, and a \textit{Multi} scene combining all features, illustrated in \cref{fig:setups}(b)-(f). \cref{fig:manipulation} demonstrates both the pre-grasping and grasping manipulations executed by the system, including the generated pre-grasping affordance maps.

\subsection{Real-world Results and Analysis}

\Cref{table:real_result} shows the outcomes from our real-world experiments. We tested ten object categories, conducting four trials per object in each scene, resulting in 20 experiments per scene. Our framework, \method, significantly enhanced the grasping success likelihood for hard-to-grasp objects across diverse categories, demonstrating its high deployability in real-world scenarios.

\begin{table}[htb!]
    \small
    \centering
    \setlength{\tabcolsep}{3pt}
    \caption{\textbf{Real-world experiment results.} Experiments were conducted twice for each object in every scene, comparing direct grasping (without pre-grasping) to grasping after pre-grasping. Success rates are presented as percentages.}
    \label{table:real_result}
    \resizebox{\linewidth}{!}{%
        \begin{tabular}{@{}ccccccccccccc@{}}
            \toprule
            \multirow{2}{*}{\textbf{Setting}} & \multicolumn{6}{c}{\textbf{Seen categories}} & \multicolumn{6}{c}{\textbf{Unseen categories}} \\ 
             \cmidrule(lr){2-7} \cmidrule(lr){8-13}
             & edge & wall & slope & slot & multi & avg. & edge & wall & slope & slot & multi & avg.\\ 
            \midrule
            W/o pre-grasping & 0 & 0 & 0 & 0 & 0 & 0& 10 & 0 & 5 & 0 & 0 & 3\\
            With pre-grasping & 70 & 45 & 80 & 90 & 85 & 74 & 80 & 30 & 75 & 90 & 85 & 72\\
            \bottomrule
        \end{tabular}%
    }%
\end{table}

\section{Conclusion and Limitation}\label{sec:conclusion}

In this paper, we introduced a novel two-stage affordance learning framework that excels in adaptability across various object-environment configurations, shows compatibility with graspable objects, and proves deployable through rigorous real-world testing. This framework has been validated in simulation and real-world settings, underscoring its effectiveness. However, our method still faces challenges, particularly in handling objects with highly irregular shapes or extremely dynamic environments. Future work will focus on enhancing the robustness and flexibility of our approach to address these issues.

\bibliographystyle{ieeetr}
\balance
\bibliography{reference_header_shorter,reference}

\end{document}